\documentclass{article}

\PassOptionsToPackage{numbers, sort&compress}{natbib}
\usepackage[preprint]{neurips_2024}

\usepackage[utf8]{inputenc} % allow utf-8 input
\usepackage[T1]{fontenc}    % use 8-bit T1 fonts
\usepackage{hyperref}       % hyperlinks
\usepackage{url}            % simple URL typesetting
\usepackage{booktabs}       % professional-quality tables
\usepackage{amsfonts}       % blackboard math symbols
\usepackage{nicefrac}       % compact symbols for 1/2, etc.
\usepackage{microtype}      % microtypography
\usepackage{xcolor}         % colors
\usepackage{amsmath}
\usepackage{graphicx}

\usepackage{float}
\usepackage{wrapfig}

\title{Learning Locally Adaptive Metrics that Enhance Structural Representation with LAMINAR}

\author{%
  \href{mailto:christian.kleiber@stud.uni-heidelberg.de}{Christian Kleiber}{\rm ,} \href{mailto:william.oliver@iwr.uni-heidelberg.de}{William H. Oliver} {\rm and} \href{mailto:tobias.buck@iwr.uni-heidelberg.de}{Tobias Buck} \\
  Interdisciplinary Center for Scientific Computing, University of Heidelberg \\
  Im Neuenheimer Feld 205, D-69120 Heidelberg, Germany \\
  \vspace{\baselineskip} % so that removing [final] before submitting doesn't change the paper length
}

\begin{document}

\maketitle

\begin{abstract}
We present \texttt{LAMINAR}, a novel unsupervised machine learning pipeline designed to enhance the representation of structure within data via producing a more-informative distance metric. Analysis methods in the physical sciences often rely on standard metrics to define geometric relationships in data, which may fail to capture the underlying structure of complex data sets. \texttt{LAMINAR} addresses this by using a continuous-normalising-flow and inverse-transform-sampling to define a Riemannian manifold in the data space without the need for the user to specify a metric over the data a-priori. The result is a locally-adaptive-metric that produces structurally-informative density-based distances. We demonstrate the utility of \texttt{LAMINAR} by comparing its output to the Euclidean metric for structured data sets.
\end{abstract}

% Main body
\section{Motivation and related work}

Much of the analysis performed within the physical sciences depends heavily upon the geometric properties of the data used. Typically it is assumed that the Euclidean metric is best to describe this geometry. However, in many settings more information can be gleaned from the distribution of the data itself. This process is often referred to as metric learning and generally aims to define a Riemannian metric in the data space such that the structure of the data set is respected and better preserved within downstream analyses. Compared to a global metric, this would suggest that the distances resulting from a more-informative metric should be smaller between data points that exist within the same mode of the data and larger for points belonging to separate modes. It then follows that a density-based metric could be more-informative.

The idea of density-based metrics is not new. \citet{Bousquet2003} posit that a density-based metric can be used to acquire more-informative distances for semi-supervised learning. This is extended by \citet{Sajama2005}, who find geodesic shortest-paths in a kernel density weighted nearest neighbour graph. The work has since continued with density approximations \citep{Bijral2012}, proofs of convergences \citep{Hwang2016, Chu2020}, and practical implementations \citep{Little2022, Groisman2022, Trillos2023}. A recent work \citep{Sorrenson2024} has shown that although proofs of convergences exist for these implementations, in practice these methods do not generally converge to a ground-truth density-based geodesic. This same work also suggests a new normalising-flow- and score-model-based method of finding a density-based metric, which is shown to converge in practice.

The irony of these approaches and of the nature of defining a more-informative metric is that a meaningful metric is typically already needed to define the meaning of \textit{local} and therefore density. We are only aware of one such class of approaches \citep{Sharma2006, Sharma2009} that does not assume a metric a-priori -- which use a scale-invariant entropy-based method to define a meaningful locally-adaptive-metric for the purpose of enhancing structural representation in cosmological simulations. In this work, we seek to improve upon these implementations by utilising inverse-transform-sampling and continuous-normalising-flows to develop an unsupervised locally-adaptive-metric (LAM) algorithm that enhances structural representation and that may be applied within the broader physical sciences.

\section{LAMINAR}

Our approach finds \textbf{L}ocally \textbf{A}daptive \textbf{M}etrics using \textbf{I}nvertible \textbf{N}etworks on \textbf{A}nisotropic \textbf{R}egions (\texttt{LAMINAR}) and calculates structurally-representative distances between any two points from an input data set. To do this, \texttt{LAMINAR} first transports the $d$-dimensional input data to a uniform distribution within the volume of the $d$-dimensional unit sphere. This transformation behaves like a cumulative distribution function (i.e. a pseudo-cdf) in the sense that it is uniform and that locality is preserved. Therefore, by then defining a Euclidean $k$-nearest-neighbour graph in the pseudo-cdf, \texttt{LAMINAR} attains an inverse-transform sampling that connects each point to a structurally-representative neighbourhood of the data space. \texttt{LAMINAR} then computes neighbour-neighbour distances (edge-weights) with a LAM (where the metric tensor varies according to the Jacobian of the data $\mapsto$ pseudo-cdf transformation) and longer-range distances as the sum of edge-weights belonging to shortest-paths within the graph. This process is unsupervised, density-based, and does not require a metric to be defined over the data space a-priori. The following subsections contain additional details of the implementation.

\paragraph{Transforming the input data}
The core of \texttt{LAMINAR} consists of a continuous planar flow as introduced in \cite{Chen2019}. Briefly summarised, a (discrete) planar flow is described by 
\begin{equation}
    \bold{z}(t+1) = \bold{z}(t) + uh(w^\text{T}\bold{z}(t)+b),~~ \log{(p(\bold{z}(t+1)))} = \log{(p(\bold{z}(t)))} - \log{\bigg|1+u^\text{T}\frac{\partial h}{\partial \bold{z}}\bigg|}.
\end{equation}
Note that $u$, $w$ $\in \mathbb{R}^d$ and $b$ $\in \mathbb{R}$ are adjustable parameters, $h$ an activation function, $t$ describes an arbitrary time scale, and $z(t)$ is a random variable distributed according to $p(z(t))$. Using the parallels of this ResNet structure to the Euler discretisation of continuous transformations, and the instantaneous change of variables, the continuous planar flow is then given by
\begin{equation}
    \frac{d\bold{z}(t)}{dt} = uh(w^\text{T}\bold{z}(t)+b),~~ \frac{\partial\log{(p(\bold{z}(t)))}}{\partial t} = - u^\text{T}\frac{\partial h}{\partial \bold{z}(t)}.
\end{equation}
Here, $h$ is an activation function adhering to Lipschitz continuity and $u$, $w$ and $b$ are time-dependent, learnable parameters, given as an output of a hypernetwork (a fully connected MLP with a single hidden layer and only time as an input). This combined ODE can be solved using the given initial distribution, $p(\bold{z}(0))$, and since the result, $p(\bold{z}(t))$, can be evaluated as having been drawn from a probability distribution -- as such the training is guided by the log-likelihood loss. Once \texttt{LAMINAR} has finished the training, the ODE will describe a continuous transformation from any $d$-dimensional initial distribution to a $d$-dimensional standard-normal distribution. \texttt{LAMINAR} then further transforms this multivariate Gaussian into a uniform distribution within the $d$-dimensional unit sphere, by adjusting the radius of each point according to
\begin{equation} \label{eq:gausstosphere}
    \bold{r}_\mathrm{sphere} = \frac{\bold{r}_\mathrm{gaussian}}{|\bold{r}_\mathrm{gaussian}|} \cdot F(\bold{r}_\mathrm{gaussian}),\ \mathrm{with}\ F(\bold{r}) = \bigg(1 - \frac{\Gamma(\frac d2, \frac{\bold{r}^2}{2})}{\Gamma(\frac d2)}\bigg)^{\frac 1d},
\end{equation}
where $F(\cdot)$ is the CDF of the multivariate Gaussian and $\Gamma(\cdot, \cdot)$ is the upper incomplete Gamma function.
%
%\begin{equation}
%    F(\bold{r}) = \bigg(1 - \frac{\Gamma(\frac d2, \frac{\bold{r}^2}{2})}{\Gamma(\frac d2)}\bigg)^{\frac 1d}
%\end{equation}
%
%with $\Gamma(\cdot, \cdot)$ being the upper incomplete Gamma function.

%With this subsequent transformation, the flow model transforms any $d$-dimensional distribution into a uniform distribution within the volume of a $d$-dimensional unit sphere. \red{In this latent space, every dimension spans the maximal range from $-1$ to $1$ and there are no significant local density variations anymore, justifying the use of the Euclidean metric for distance calculations. \#This does sound weird, but not sure how to phrase that differently\#} % I added text at the start of this section which I think removes the need for this here.

\paragraph{Calculating distances}
Once \texttt{LAMINAR} has transformed the data, the $k$-nearest-neighbours of each point, $N_k(\bold{x}_i)$, are found from the pseudo-cdf using a KDTree and a Euclidean-metric. These neighbours can be understood as defining connections in an adjacency matrix in the original data space, whose weights are Mahalanobis distances (analogous to \citet{Sharma2009}):
%
%At this point, a grid is used for an additional approximation of distances, meaning that the minimum and maximum ranges of each dimension in the data space are used as boundaries for a $d$-dimensional grid. This grid is also pushed through the model and the results are concatenated with the pushed data. In the next step, each pushed point gets assigned its $k$-nearest neighbours in the latent space via KDTrees. These neighbours can be understood as connections in an adjacency matrix, whose weights are defined as Mahalanobis distances, analogous to \citet{Sharma2009}:
%
\begin{equation} \label{eq:mahalanobis}
    s^2(\bold{x}_i, \bold{x}_j) = |\bold{\Sigma}(\bold{x}_i, \bold{x}_j)|^{1/d}\cdot(\bold{x}_i - \bold{x}_j)^{\text{T}}\cdot\bold{\Sigma}(\bold{x}_i, \bold{x}_j)^{-1}\cdot(\bold{x}_i -\bold{x}_j),
\end{equation}
where $\bold{\Sigma}(\bold{x}_i, \bold{x}_j) = 0.5\left[\bold{\Sigma}(\bold{x}_i)) + \bold{\Sigma}(\bold{x}_j)\right]$
%
%\begin{equation}
%    \bold{\Sigma}(\bold{x}_i, \bold{x}_j) = 0.5 \cdot (\bold{\Sigma}(\bold{x}_i)) + \bold{\Sigma}(\bold{x}_j))
%\end{equation}
%
is the mean metric tensor of the two queried points e.g. neighbours in the adjacency matrix. Here, $\bold{\Sigma}$ is defined using the Jacobian of the full transformation such that
\begin{equation} \label{eq:metrictensor}
    \bold{\Sigma} = (\bold{J}_\mathrm{total}^\text{T} \cdot \bold{J}_\mathrm{total})^{-1},\ \mathrm{with}\ \bold{J}_\mathrm{total} = \bold{J}_\mathrm{to\ sphere} \cdot \bold{J}_\mathrm{flow}.
\end{equation}
For completeness, $\bold{J}_\mathrm{flow}$ is calculated using PyTorch's autograd function, while $\bold{J}_\mathrm{to\ sphere}$ is calculated analytically using Eq.~\ref{eq:gausstosphere} -- and both are evaluated at the position of each point, $\bold{x}_i$.
\texttt{LAMINAR} can then calculate the distance between any two points of the input data, $d_{ij}$, via Dijkstras's algorithm \citep{Dijkstra1959} -- which searches for the shortest-path within the graph defined by the aforementioned adjacency matrix filled with the new local distances as in Eq. \ref{eq:mahalanobis}. For clarity; given any two points, e.g. $\bold{x}_i$ and $\bold{x}_j$, if $\bold{x}_i \in N_k(\bold{x}_j)$ or $\bold{x}_j \in N_k(\bold{x}_i)$, then $d_{ij} = s(\bold{x}_i, \bold{x}_j)$ from Eq.~\ref{eq:mahalanobis}.

\section{Visualisation of the LAMINAR metric}
It is difficult to quantitatively assess the \texttt{LAMINAR}'s performance since this requires measuring the degree of structural meaningfulness produced in the resultant metric. Nevertheless, we can assess its ability to match a set of \textit{ground-truth} metrics determined analytically from simple transformations of a uniform circle (where the Euclidean metric is structurally meaningful). Although, we don't necessarily expect \texttt{LAMINAR} to match these metrics perfectly as there are an infinity of ways that each data point could be transformed while still producing the same final distribution (consider applying transformation $T(\bf{x})$ vs. applying $T(\bf{R}\bf{x})$ where $\bf{R}$ is some rotation). Hence with this analysis we assess whether LAMINAR does what we have claimed it to do and begin to demystify its behaviour.

\begin{figure}[ht]
        \vspace{-0.25cm}
        \centering
	\includegraphics[width=0.95\textwidth]{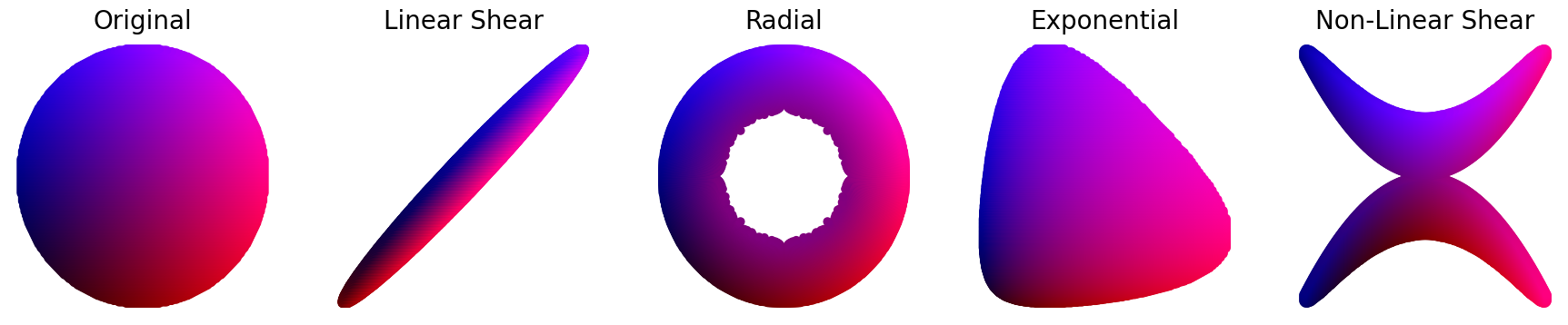}
        \vspace{-0.25cm}
	\caption{The original (uniform) distribution and the transformations applied to it.}
	\label{fig:transformations_new}
        \vspace{-0.25cm}
\end{figure}

\begin{figure}[ht]
        \vspace{-0.25cm}
        \centering
	\includegraphics[width=0.95\textwidth]{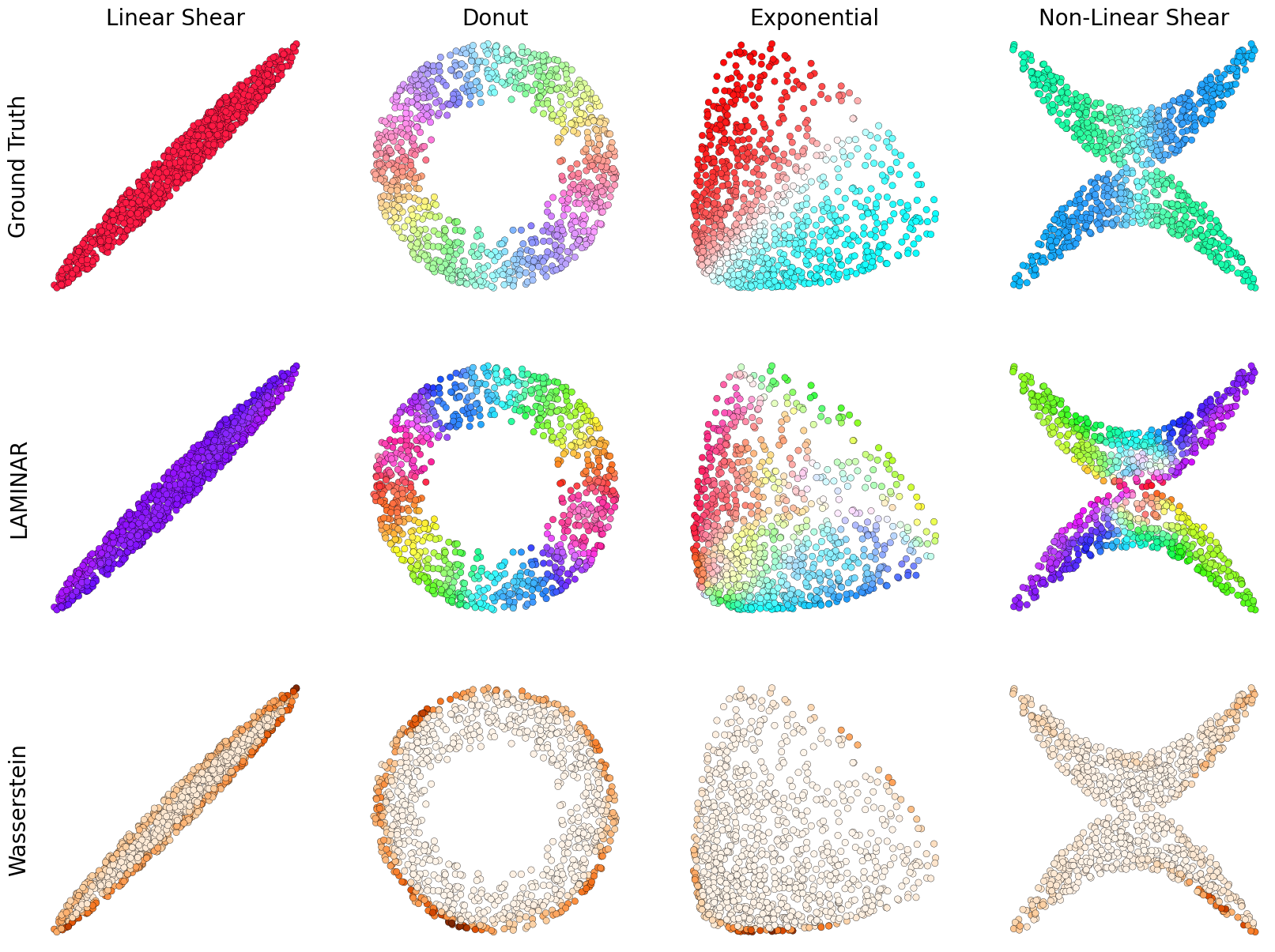}
	\vspace{-0.25cm}
	\caption{Comparison of \textit{ground-truth} and \texttt{LAMINAR} metric tensors produced using data in Fig. \ref{fig:transformations_new}.}
	\label{fig:metric_vis}
        \vspace{-0.25cm}
\end{figure}

\begin{figure}
    %\vspace{-0.25cm}
    \begin{minipage}[c]{0.2\textwidth}
        \vspace{0.2cm}
        \includegraphics[width=\textwidth]{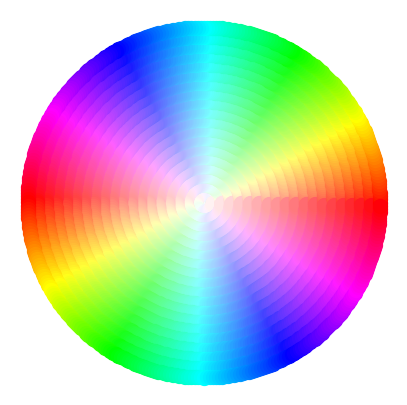}
    \end{minipage}\hfill
    \begin{minipage}[c]{0.8\textwidth}
        \caption{A reference colour wheel for the visualisation of the metric tensor in Fig. \ref{fig:metric_vis}. To assign a colour to a data point, we first create an ellipse by transforming a circle with that point's metric tensor. The colour (angle) assigned is given by the orientation of this ellipse, i.e. red (blue) if its major axis aligns horizontally (vertically). The saturation (radius) of this colour is determined by the ratio between the lengths of the major and minor axes -- so that a more spherical ellipse is lighter in colour. Visualising the metric in this way shows us the direction in, and degree to, which the distance function increases most (from each data point).} \label{fig:reference}
    \end{minipage}
    \vspace{-0.5cm}
\end{figure}

Fig. \ref{fig:transformations_new} depicts four transformations while for each, Fig. \ref{fig:metric_vis} shows the ground-truth metric, the metric predicted by \texttt{LAMINAR}, and the point-wise difference between these metrics (measured with the Wasserstein distance between multivariate normal distributions whose covariance matrices are the metric tensors). The metrics are visualised according to the colour scheme described with Fig. \ref{fig:reference}.

%\begin{wrapfigure}[12]{R}{0.3\textwidth}
%        \vspace{-0.25cm}
%        \centering
%	\includegraphics[width=0.29\textwidth]{Figures/reference.png}
%        \vspace{-0.25cm}
%	\caption{Reference colour wheel for the orientation of ellipses.}
%	\label{fig:reference}
%        \vspace{-0.25cm}
%\end{wrapfigure}

In Fig. \ref{fig:metric_vis} we see that while there is some discrepancy due to noise as well as due to edge effects (related to the boundary of the pseudo-cdf), \texttt{LAMINAR} is able to learn the correct metric having only seen the transformed data set. It is worth noting that in the case of the linear shear transformation, the metric predicted by \texttt{LAMINAR} is seemingly a better representation of the global structure than that provided by the \textit{ground-truth} -- the \textit{ground-truth} here results from the data experiencing a $y$-dependent translation while the \texttt{LAMINAR} metric corresponds to the global Mahalanobis distance one can recover by calculating this data set's covariance matrix. Still, these results suggest that \texttt{LAMINAR} would likely benefit from using an optimal-transport-based flow.

%Also the differences between the metric are predominantly at the edges of the data clouds, which may come from the fact that \texttt{LAMINAR} only knows the data and has no information beyond that, therefore it cannot properly extrapolate beyond, while the known transformation is independent of the data. 

%Reproduction of GT metric would only be expected if the GT transformation were constrained to optimal transport maps, and LAMINAR was to be expanded to converge to optimal flows.  

\section{Direct comparison between the Euclidean metric and LAMINAR} \label{sec:directcomparison}
\begin{figure}[ht]
        \centering
        \vspace{-0.25cm}
	\includegraphics[width=0.95\textwidth]{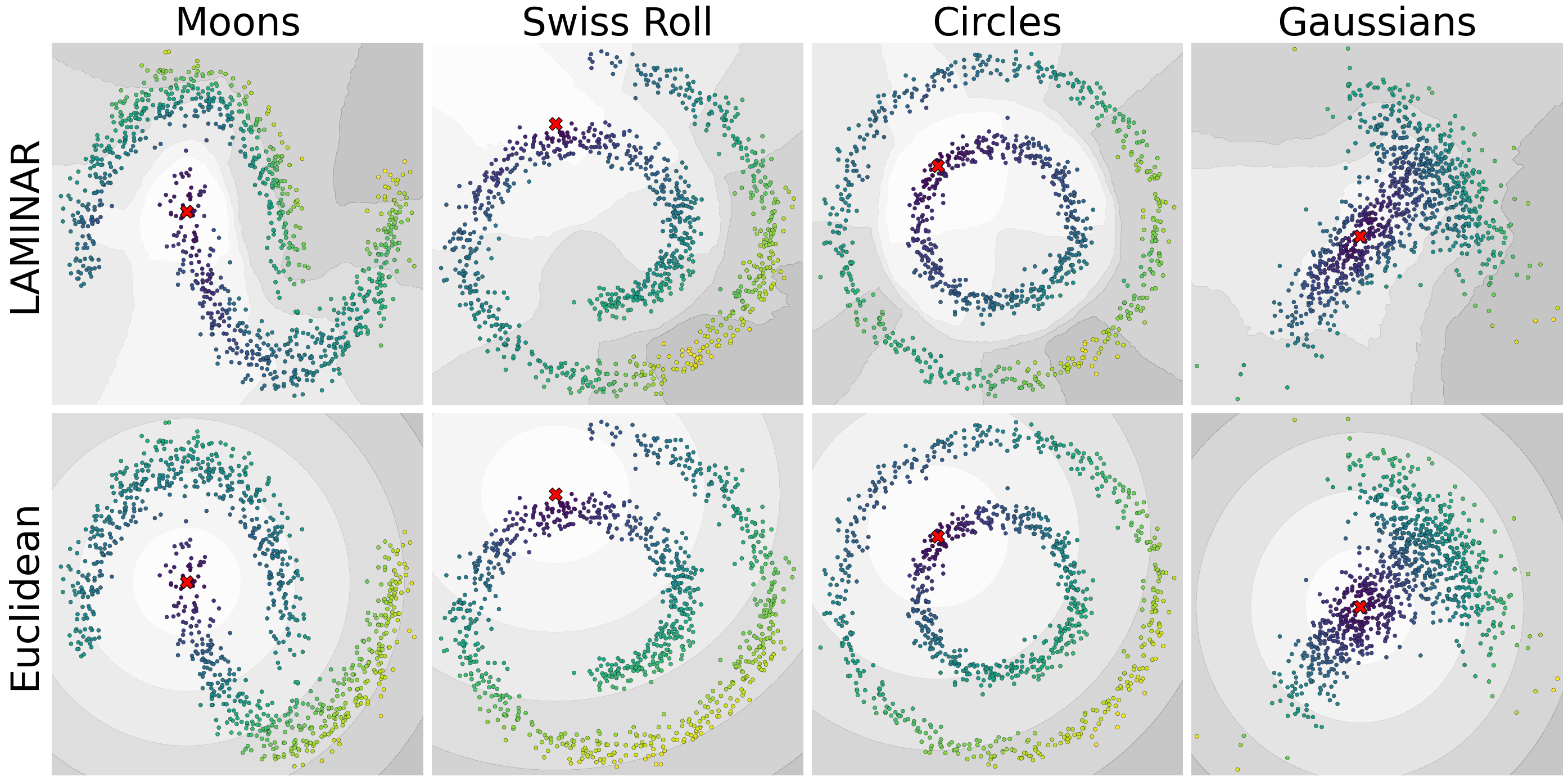}
        \vspace{-0.25cm}
	\caption{The distance distribution from a query point (red cross) on four toy data sets found with the \texttt{LAMINAR} (top) and Euclidean (bottom) metrics. The points are coloured according to the viridis colour map. The grey-scale contours show an estimate for the out-of-distribution distances.}
	\label{fig:LAMvsEuc}
        \vspace{-0.5cm}
\end{figure}

We now conduct a qualitative check to see how \texttt{LAMINAR} compares to the Euclidean metric globally. Fig. \ref{fig:LAMvsEuc} shows a few toy data sets coloured according to their distance from a query point, where brighter (darker) colours indicate smaller (larger) distances. The shortest-paths found by \texttt{LAMINAR} prefer dense regions, adapting to the data, while naturally those from the Euclidean metric are data-independent. The contours show an approximation of how the distance may look for points beyond the data set -- estimated as the average distance-value of the $k = 25$ nearest neighbours. %\red{Note that this is no accurate representation for low density regions and in these cases rather serves as an illustrative purpose.}

\begin{figure}[ht]
	\centering
        \vspace{-0.25cm}
	\includegraphics[width=0.95\textwidth]{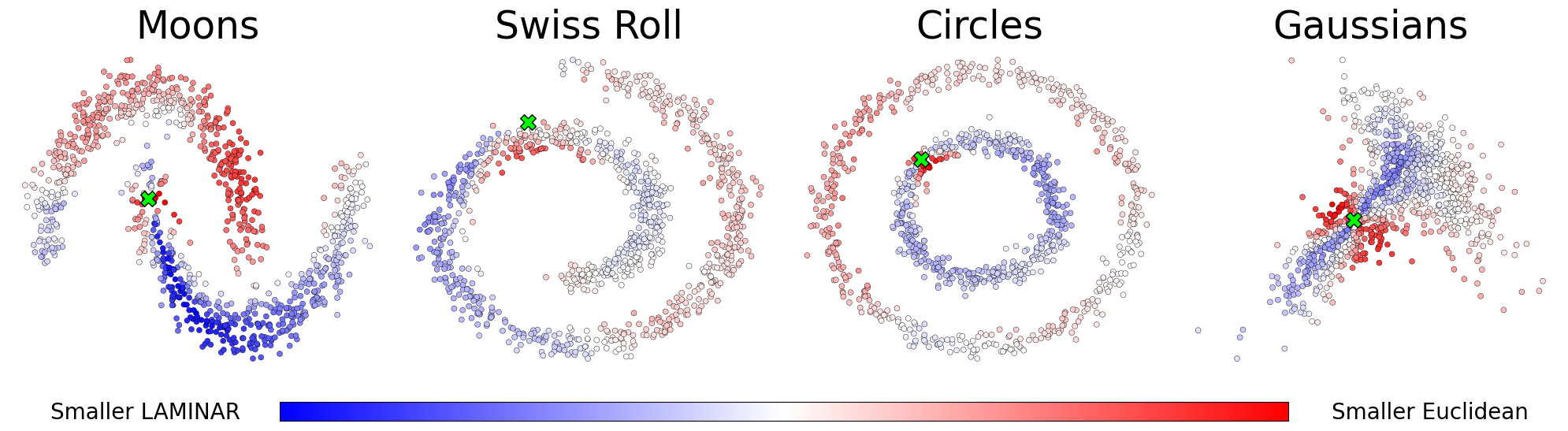}
        \vspace{-0.25cm}
	\caption{A comparison of distances from query points (marked with a green cross) produced via the \texttt{LAMINAR} and Euclidean metrics (as shown in Fig~\ref{fig:LAMvsEuc}). Here blue points mark those with smaller \texttt{LAMINAR} distances compared to their Euclidean counterparts, and \textit{vice versa} for red points.}
	\label{fig:LAMvsEuc2}
        \vspace{-0.25cm}
\end{figure}

Fig. \ref{fig:LAMvsEuc2} illustrates a more direct comparison of the Euclidean and \texttt{LAMINAR} metrics. For each metric, we take the logarithm of the distances shown in Fig. \ref{fig:LAMvsEuc} and standardise each by subtracting the mean and dividing by the standard deviation -- these values for each metric are then subtracted from one another. The resultant quantity represents a ratio of distances, and while the exact value is not meaningful, it portrays how the \texttt{LAMINAR} metric behaves relative to the Euclidean one. Again, we see that moving along the modes of the data is preferred (discouraged) by the \texttt{LAMINAR} (Euclidean) metric, while moving perpendicular to the data is treated oppositely. As such, it is clear that \texttt{LAMINAR} is able to learn and emphasise the structure implicit to the data sets.

\section{Usage in downstream analysis}

\begin{figure}[ht]
	\centering
        \vspace{-0.25cm}
	\includegraphics[width=0.95\textwidth]{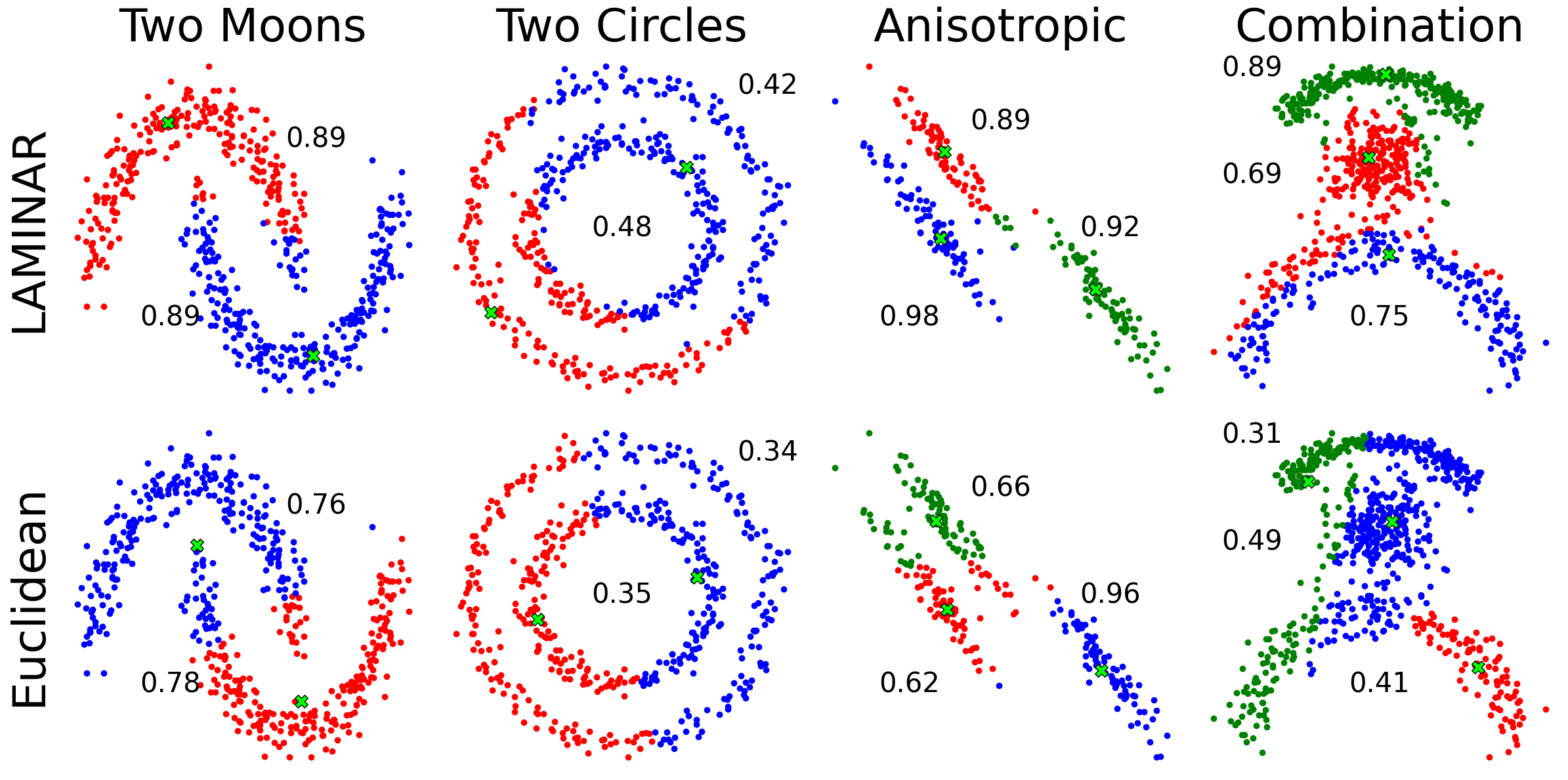}
        \vspace{-0.25cm}
	\caption{Comparison of the \texttt{LAMINAR} (top) and Euclidean (bottom) metrics in the downstream clustering tool, \texttt{k-medoids}. Values shown next to the clusters are Jaccard similarities between the ground-truth clusters and the best-fitting predicted cluster (colours).}
	\label{fig:K-Medoids}
        \vspace{-0.25cm}
\end{figure}

To demonstrate the advantage of \texttt{LAMINAR} in downstream analyses, an example is provided by using \texttt{LAMINAR}-calculated distances in the \texttt{k-medoids} algorithm for clustering. These results are compared to the same for Euclidean distances, i.e. the standard metric used in such algorithms. \texttt{k-medoids} is chosen as it allows clustering without deviating from the data points (in contrast to k-means) -- since this early version of \texttt{LAMINAR} is not able to calculate distances between out-of-distribution points and is instead limited to distances between existing data points. Additionally, \texttt{k-medoids} is a very simple algorithm, so the choice of metric has a large impact on its performance. In contrast, it is possible that other more complicated clustering algorithms may not see as much improvement due to a focus on optimizations using the standard metric. These results are shown in Fig. \ref{fig:K-Medoids}.

Even though the ground-truth clusters are visibly apparent, they pose significant problems for \texttt{k-medoids}. We assess the effect of the different metrics by calculating the Jaccard index between each of the ground-truth clusters and the best-fitting predicted cluster from each of the metrics. Here, a value of $1$ shows perfect agreement and flawless reconstruction of the cluster while $0$ shows no overlap of the two clusters. We can see from Fig. \ref{fig:K-Medoids} that the effect of the \texttt{LAMINAR} metric is to improve cluster extraction for (almost) all cases -- in certain cases almost achieving a perfect reconstruction. This demonstrates the usability of \texttt{LAMINAR} in downstream analysis for enhanced performance and results that depend on the structural representation of the data.

\section{Conclusions and outlook}

We have introduced \texttt{LAMINAR}, a novel unsupervised machine learning pipeline that generates a locally-adaptive-metric to enhance the structural representation of data. \texttt{LAMINAR} is able to compute more-informative distances that preserve the underlying global structure of the data more effectively than the traditional Euclidean metric. Our approach avoids the need for pre-defining a meaningful metric, which is often a limitation in other density-based methods, and provides a robust means of uncovering the geometry inherent in the data itself. Our current implementation, found at \url{https://github.com/CKleiber/LAMINAR}, is a proof-of-concept and in a future work we aim to incorporate the techniques developed by \citet{Sorrenson2024}, employ an optimal-transport-based flow architecture, as well as to investigate whether additional geometric information can be utilized in order to enhance structural representation implicit within a data set.

%All in all, \texttt{LAMINAR} shows its potential in preserving local neighbourhoods despite unknown transformations as good as, if not even better than the Euclidean metric. Furthermore, the qualitative tests show a preference of \texttt{LAMINAR} to follow the data dense regions, while avoiding gaps, hence incorporating more available information about the macro structure of the data distribution in its distance assessment, hinting at superiority in certain contexts over global data-independent metrics such as the Euclidean.

%While the isolated tests show promising results, the actual use for application is restricted by the long run time due to the precomputing of the Jacobians for every data point. This can be avoided by proceeding to avoid the grid and discrete calculations and implementing continuous methods with geodesics and geodesic solvers for calculating distances, similar to \citet{Sorrenson2024}. 

%This graph theoretical approach is to be understood as an intermediate step towards continuous geodetic connections, eventually also allowing the calculation of any two points in the data space and not only the actual data points. 

\section*{Broader impact statement}
The authors are not aware of any immediate ethical or societal implications of this work. This work purely aims to aid scientific research and proposes to apply normalizing flows to learn a meaningful distance metric that respects the structure implicit within any given point-based data set.

\begin{ack}
This work is funded by the Carl-Zeiss-Stiftung through the NEXUS program.
\end{ack}

\bibliography{references}

\end{document}